\documentclass[runningheads]{llncs}
\usepackage{graphicx}
\usepackage{cite} 
\usepackage{times}
\usepackage{epsfig}
\usepackage{amsmath}
\usepackage{amssymb}
\usepackage{mwe}
\usepackage{acro}
\usepackage{amssymb}
\usepackage{xcolor,colortbl}
\usepackage{tabularx}
\usepackage{relsize}
\usepackage{pifont}
\usepackage{booktabs} 
\usepackage{multirow}
\usepackage{multicol}
\usepackage{adjustbox}
\usepackage{float}
\usepackage{cuted}
\usepackage{graphicx, caption}
\usepackage[colorlinks,
            linkcolor=red,  
            anchorcolor=blue, 
            citecolor=green,   
            ]{hyperref}

\DeclareAcronym{ROI}{
short=ROI,
long=region of interest,
}

\DeclareAcronym{IOU}{
short=IOU,
long=intersection over union,
}

\DeclareAcronym{cIOU}{
short=cIOU,
long=circle intersection over union,
}

\DeclareAcronym{DoF}{
short=DoF,
long=degrees of freedom,
}

\begin{document}
\title{Contrastive Learning Meets Transfer Learning:  \\ A Case Study In Medical Image Analysis}
%
%
\author{Yuzhe Lu \and
Aadarsh Jha \and
Yuankai Huo}


%
\institute{Computer Science, Vanderbilt University, \newline Nashville TN 37235, USA}
%
\maketitle              

\begin{abstract}
Annotated medical images are typically rarer than labeled natural images, since they are limited by domain knowledge and privacy constraints. Recent advances in transfer and contrastive learning have provided effective solutions to tackle such issues from different perspectives. The state-of-the-art transfer learning (e.g., Big Transfer (BiT)) and contrastive learning (e.g., Simple Siamese Contrastive Learning (SimSiam)) approaches have been investigated independently, without considering the complementary nature of such techniques. It would be appealing to accelerate contrastive learning with transfer learning, given that slow convergence speed is a critical limitation of modern contrastive learning approaches. In this paper, we investigate the feasibility of aligning BiT with SimSiam. From empirical analyses, different normalization techniques (Group Norm in BiT vs. Batch Norm in SimSiam) is the key hurdle of adapting BiT to SimSiam. When combining BiT with SimSiam, we evaluated the performance of using BiT, SimSiam, and BiT+SimSiam on CIFAR-10 and HAM10000 datasets. The results suggest that the BiT models accelerate the convergence speed of SimSiam. When used together, the model gives superior performance over both of its counterparts. We hope this study will motivate researchers to revisit the task of aggregating big pretrained models with contrastive learning models for image analysis.


\end{abstract}
\section{Introduction}
Large-scale annotated medical images are typically more difficult to achieve compared with natural images, limited by domain knowledge and potential privacy constraints~\cite{willemink2020preparing,huo2021ai}. As a result, numerous efforts have been made to develop  a more effective image analysis algorithm with a limited number of annotated data~\cite{razzak2018deep}. On the other hand, large-scale unlabeled data can be available from the Picture Archiving and Communication System (PACS) in hospitals~\cite{hosny2018artificial}. Herein, it is appealing to obtain a good feature representation using unsupervised learning and large-scale unannotated data, and then fine-tune the learned model to downstream tasks~\cite{shen2017deep}. 

Recently, contrastive learning techniques — a new cohort of unsupervised representation learning — demonstrated its superior performance in various vision tasks~\cite{wu2018unsupervised,noroozi2016unsupervised,zhuang2019local,hjelm2018learning}. The recent studies showed that fine-tuning on a well trained contrastive learning model leads to comparable, sometimes even better results, compared with fully supervised learning. SimCLR~\cite{chen2020simple} was a pioneer study in contrastive learning, which maximized the similarity between positive pairs and the differences between negative pairs. MoCo~\cite{he2020momentum} introduced a momentum design to maintain a negative sample pool instead of an offline dictionary. More recently, BYOL \cite{grill2020bootstrap} was proposed to train a model without negative samples. Then, SimSiam~\cite{chen2020exploring} was proposed to further eliminate the momentum encoder in BYOL, for a more simpler and efficient design.

On the other hand, big transfer learning leverages vision tasks with limited data from a different perspective. ``Big Transfer'' (BiT) is one of the largest and released pretrained model~\cite{kolesnikov2019big}, which has achieved state-of-the-art results on many vision benchmarks. However, the effectiveness of big transfer learning is mostly validated on natural image datasets. Interestingly, recent studies provided contradictory conclusions of the value of transfer learning for medical imaging. Some studies presented that the value of transfer learning is limited~\cite{raghu2019transfusion,alzubaidi2020towards} , while a more recent work~\cite{mustafa2021supervised} demonstrated that pretraining at a larger scale is a key to performance gains. However, both parties agree that transfer learning speeds up the convergence of downstream tasks when being used to initialize the pretrained models. Given that slow convergence speed is one of the key limitations of modern contrastive learning approaches~\cite{chen2020simple,he2020momentum,grill2020bootstrap,chen2020exploring}, it would be appealing to accelerate contrastive learning with transfer learning.

\begin{figure}[t]
\begin{center}
\includegraphics[width=1\linewidth]{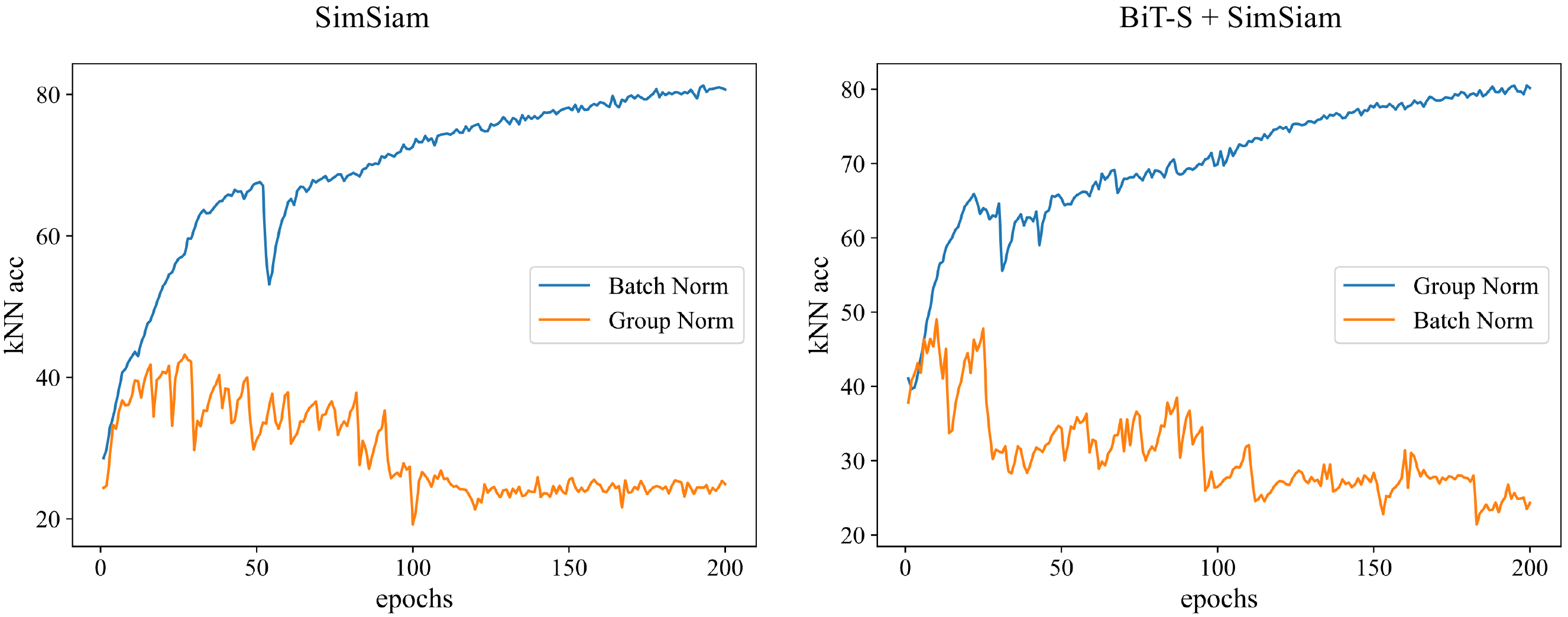}

\end{center}
   \caption{\textbf{Model collapse of SimSiam and BiT-S+SimSiam with Group Norm.} This figure shows that both SimSiam and BiT-S+SimSiam (SimSiam using BiT as the pretrain weights) encounter model collapse using Group Norm. Batch Norm is a key factor to prevent model collapse when linking BiT with SimSiam.}
\label{fig:collapse}

\end{figure}

In this paper, we investigate the feasibility of linking big transfer learning models with contrastive learning in order to accelerate contrastive learning. As a pilot study, BiT and SimSiam are employed in this study as the two representative approaches from both families. The motivations of this study are in four-fold: (1) The slow convergence speed is currently a bottleneck of contrastive unsupervised learning; (2) Big transfer learning has been validated to accelerate downstream learning tasks; (3) Such two approaches might naturally be complementary since a good common visual representation is provided by big pretrained models, while the contrastive learning would adapt the representation to specific medical image tasks from unlabeled data; and (4) both methods shared similar backbone networks (e.g., ResNet~\cite{he2016deep}), which are mutually adaptable. 

We performed a small-scale pilot study with skin lesion classification by first evaluating BiT and SimSiam. Then, we linked BiT and SimSiam as a uniformed solution (BiT+SimSiam). CIFAR-10~\cite{cifar100} and HAM10000 datasets~\cite{tschandl2018ham10000} were used as the testing data. From empirical analysis, the distinct normalization techniques (group normalization (Group Norm~\cite{wu2018group}) vs. batch normalization (Batch Norm~\cite{ioffe2015batch})) is the key hurdle of adapting the BiT to SimSiam. After replacing the Group Norm with Batch Norm while keeping the weights of remaining layers, the BiT models speed up the SimSiam learning, alongside superior accuracy. We hope this study will motivate researchers to develop a principle way of aggregating pretrained models and contrastive learning.  

\section{Methods}

\subsection{BiT and SimSiam}
To evaluate the transfer learning performance, we leverage the family of Big Transfer Models~\cite{kolesnikov2019big} that are pre-trained on natural image datasets of different scales. They utilize the ResNet-V2 \cite{he2016identity} architecture of varying sizes and substitute Batch Norm with a combination of Group Norm and Weight Standardization\cite{qiao2019weight} to maximize the benefits of transfer learning. Recent work has shown that these models are highly successful across different computer vision and medical imaging tasks. Currently, models that are pre-trained on ILSVRC-2012\cite{russakovsky2015imagenet} which contains 1.3M images (BiT-S) and ImageNet-21k\cite{deng2009imagenet} which contains 14M images (BiT-M) are released for public use. Contrastive learning methods aim to maximize the similarity between positive pairs and the dissimilarity between negative pairs. SimSiam loosens such requirements by only requiring positive samples, without needing large batch sizes for negative samples.  

\subsection{Batch Norm vs. Group Norm}
Kolesnikov et al.~\cite{kolesnikov2019big} demonstrated that Batch Norm (BN) is detrimental to Big Transfer learning due to two reasons:  ``First, when training large models with small per-device batches, BN performs poorly or incurs inter-device synchronization cost. Second, due to the requirement to update running statistics, BN is detrimental for transfer."~\cite{kolesnikov2019big}. As a result, Group Norm was employed as a standard operation for Big Transfer (BiT). On the other hand, Chen et al. elucidated that Batch Norm is a beneficial piece of SimSiam~\cite{chen2020simple}, without explicitly evaluating the performance of Group Norm. 

To combine BiT with SimSiam, the naive idea is to perform continuous contrastive learning using the BiT pretrained ResNet50-V2 backbone. However, the model collapses at the contrastive learning stage using this approach (Fig.~\ref{fig:collapse}). 


\subsection{Combine BiT with SimSiam}

Since both transfer and contrastive learning are powerful tools to alleviate the problem of lacking high-quality labels, it is natural to ask the question — can the two frameworks be combined? More specifically, transfer learning is able to not only boost model performance, but also accelerates the rate of convergence. By contrast, contrastive learning methods usually require a large number of epochs for the model to converge and provide good performance. In this study, we propose a simple way to incorporate BiT models into the SimSiam network training — simply reusing pre-trained weights of convolutional layers while replacing the Group Norm with Batch Norm via default initialization. In the rest of the paper, we will use BiT-S+SimSiam to denote training SimSiam network with pretrained weights from the BiT-S model, and BiT-M+SimSiam to denote training SimSiam network with pretrained weights from the BiT-M model.

\section{Data Details}

\indent\indent\textbf{CIFAR-10.} The CIFAR-10 dataset contains 60,000 32x32 natural images with 10 classes~\cite{cifar100}. Previous contrastive learning methods have achieved highly accurate predictions on this dataset, and we utilize such a dataset mainly to illustrate the importance of Batch Norm in the encoder of transferring BiT models to SimSiam contrastive learning.

\textbf{HAM10000} The medical image data used in this study is HAM10000, a publicly available dataset that consists of 10,015 dermatoscopic images from seven classes~\cite{tschandl2018ham10000}. Like most medical imaging datasets, HAM10000 is also highly imbalanced, where 67\% of the data comes from nevi, 11\% comes from melanoma, and the remaining 22\% comes from the other five classes. In our experiments, we randomly sampled 60\% of the dataset for unsupervised pretraining, 20\% of the dataset for linear classifier fine-tuning and validation, and 20\% of the dataset for testing. To summarize, only 10\% of the labels from HAM10000 were used for training the network during the fine-tuning stage. All the data is resized to $224$x$224$ in our experiments. 

\section{Experiments}
\subsection{Baseline Settings}
In all of our contrastive learning experiments, we use the SimSiam framework with the classic ResNet50-V2 architecture as the backbone. We use the SGD optimizer with weight decay set to $0.0005$, and momentum set to $0.9$. We use a base learning rate of $0.03$ and follow a linear scaling($lr\times$BatchSize/256)\cite{goyal2017accurate}. We also apply a cosine decay schedule~\cite{chen2020exploring} on the learning rate. We use a 128 Batch Size, which achieves decent performance in SimSiam~\cite{chen2020simple}. With mixed precision training~\cite{micikevicius2018mixed}, we are able to fit the training into a single 16GB memory GPU. 

\subsection{CIFAR-10 Experiments}
Since images from CIFAR-10 are relatively small, we change the ResNet50-V2's convolutional head from $7\times7$ with stride $2$ to $3\times3$ with stride $1$ and remove the max pooling layer as in \cite{chen2020simple}. We train two sets of contrastive learning models on CIFAR-10 to demonstrate the importance of Batch Norm on the backbone, with one set loading the pretrained BiT-S weights and the other trained from scratch. Within each set, we train one model with the Group Norm and Weight Standardization combination originally implemented in the BiT models, and train the other model by replacing the Group Norm with Batch Norm using the default PyTorch initialization. 

\subsection{HAM10000 Experiments}
On the HAM10000 dataset, we investigate the impact of supervised pre-trained weights on contrastive learning. In all experiments on HAM10000, we modified the ResNet50-V2 architecture from BiT by substituting Group Norm for Batch Norm with the default PyTorch initialization and loading the pre-trained weights for convolution layers. We trained for 400 epochs for all experiments on HAM10000.

\begin{figure}[t]
\begin{center}
\includegraphics[width=1\linewidth]{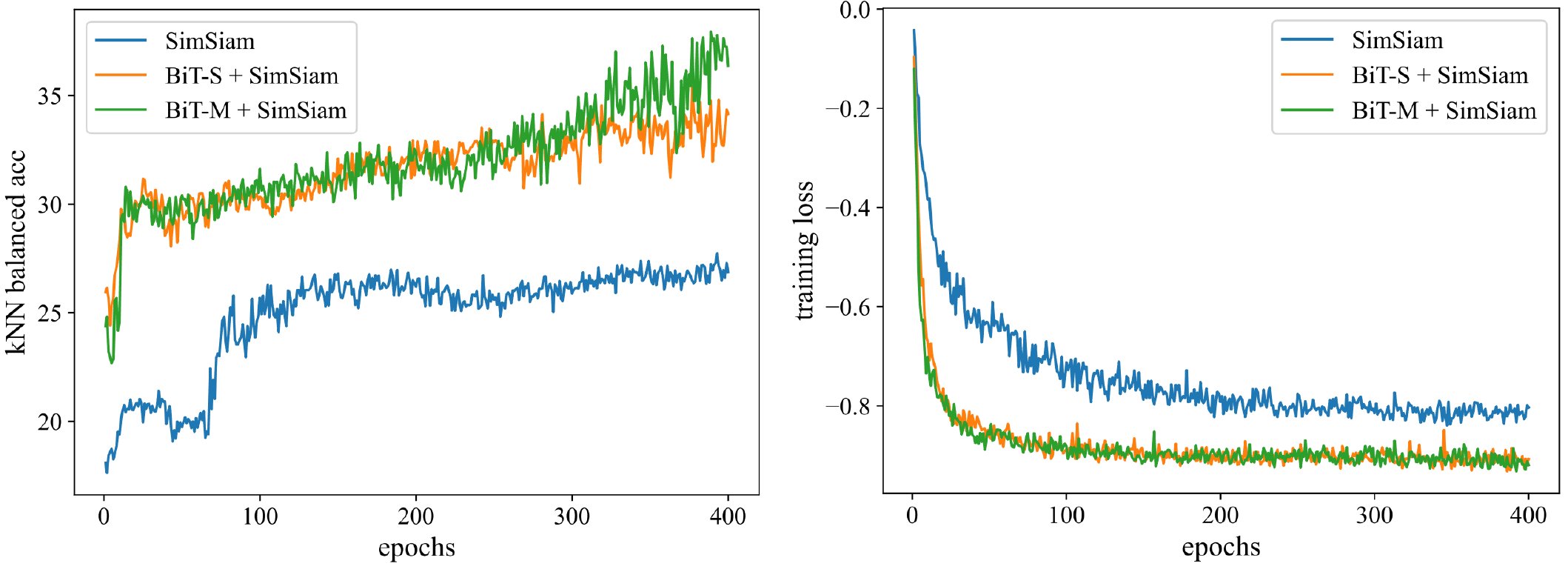}

\end{center}
\caption{\textbf{Training performance of SimSiam with BiT on HAM10000.} The left panel shows the results of the k-nearest-neighbor\cite{wu2018unsupervised} balanced accuracy (kNN balanced acc), while the right panel shows the training loss curves.}
\label{fig:ham_train}
\end{figure}

\begin{figure}[t]
\begin{center}
\includegraphics[width=0.7\linewidth]{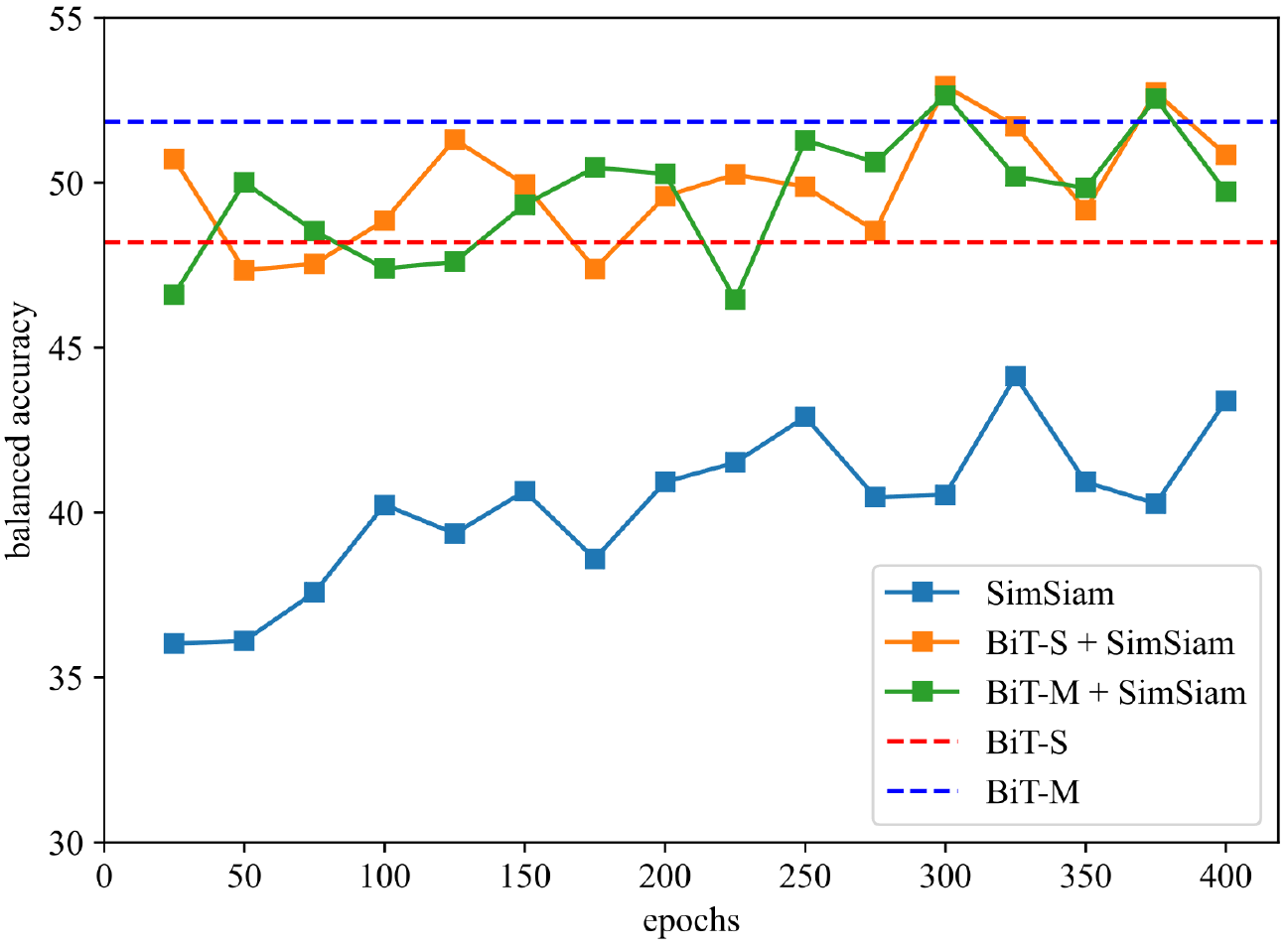}
\end{center}
   \caption{\textbf{Testing performance of SimSiam with BiT on HAM10000.} This plot shows the testing performance on HAM10000 dataset by fine tuning different learning models. }
\label{fig:ham_exp}
 \end{figure}

\begin{figure}[t]
\begin{center}
\includegraphics[width=1\linewidth]{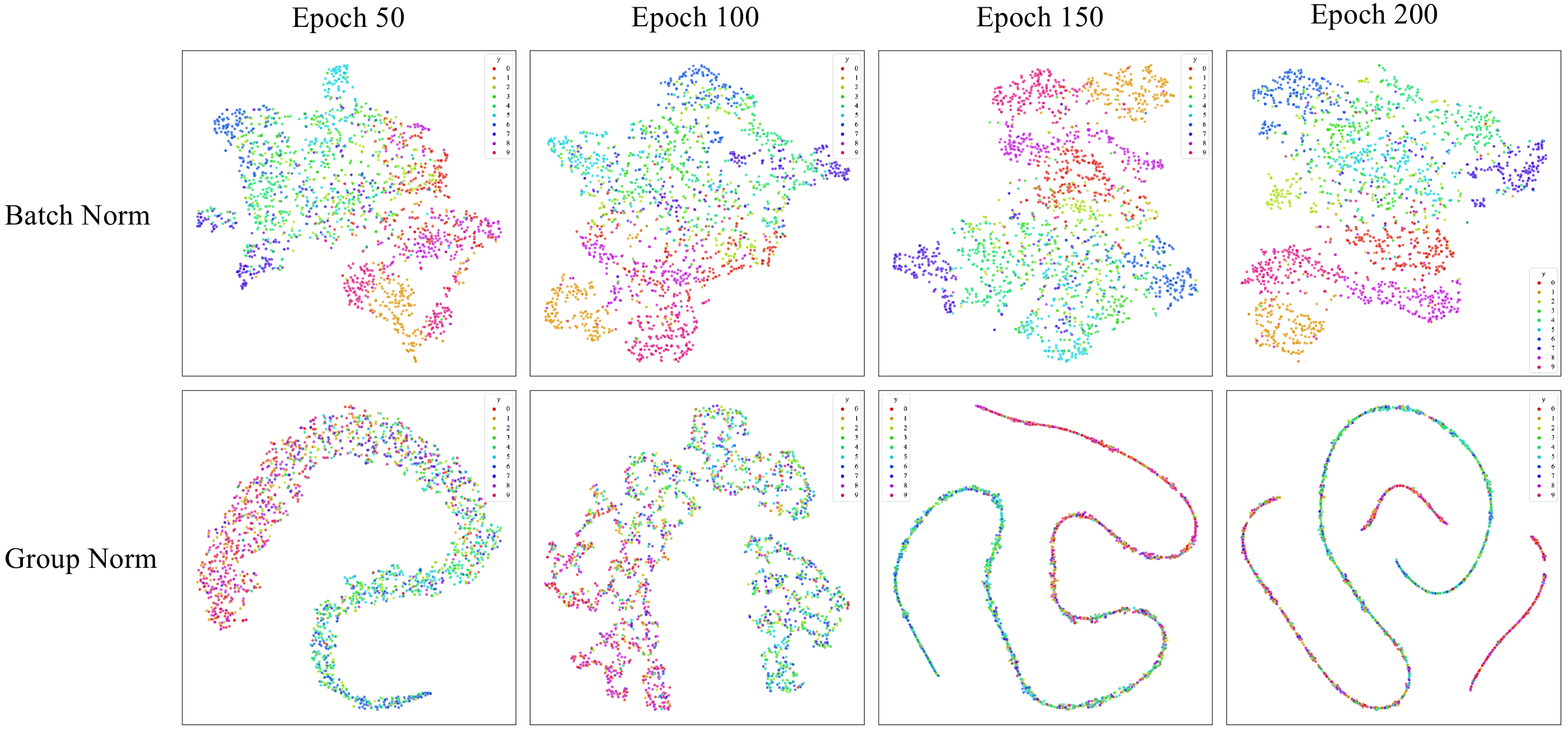}
\end{center}
   \caption{\textbf{The t-SNE plots of SimSiam with Batch Norm and Group Norm.} The features learned by SimSiam on CIFAR-10 are presented. The feature embedding collapsed when Group Norm is used for SimSiam.}
\label{fig:cifar_tsne}
\end{figure}

\subsection{Linear Evaluation}
The linear evaluation protocol is used in our study to compare the learned representations on HAM10000. For both BiT and the trained contrastive learning model backbones, we freeze all layers of the model and train a linear classifier with 10\% of the data reserved for fine-tuning. Due to the imbalanced nature of the dataset, we use focal loss with $\gamma = 4$. Furthermore, we use the balanced accuracy score as the main evaluation metric. The linear classifier with the best balanced accuracy score on validation set is used for testing. The balanced F1 score is also calculated as a supplementary evaluation metric. To reduce the variance of the reported results, we perform $3$ trials of fine-tuning on each model and report the average performance.

\section{Results}
\subsection{BiT to SimSiam}
Fig.~\ref{fig:collapse} shows the results of adapting the pretrained ResNet-50 model to SimSiam on CIFAR-10. The result demonstrates that directly applying the pretrained model as the encoder would yield a model collapse (see training accuracy) with Group Norm. Fig.~\ref{fig:ham_train} shows that by using pretrained models and replacing Group Norm with Batch Norm, contrastive learning is able to perform reasonable learning behaviors with transfer learning on HAM10000.

\subsection{BiT vs. SimSiam}
Fig.~\ref{fig:ham_exp} shows the results of five strategies: BiT-S, BiT-M, SimSiam, BiT-S+SimSiam, and BiT-M+SimSiam. For BiT, both BiT-S and BiT-M are employed to evaluate the impacts of transfer learning with different sizes of training data. From the results, BiT-M achieves better balance acc compared with BiT-S. BiT-S+SimSiam uses the BiT-S pretrained model as weight initialization, while BiT-M+SimSiam utilizes the BiT-M model as weight initialization. Fig.~\ref{fig:ham_train} and~\ref{fig:ham_exp} demonstrate that BiT speeds up the convergence of SimSiam, when comparing to BiT-S+SimSiam, BiT-M+SimSiam and SimSiam. Moreover, the performance of initialization by the BiT pretrained model is generally better than training SimSiam from scratch. BiT-S+SimSiam outperforms BiT-S, while achieving comparable performance to BiT-M.

\subsection{Visualization on Feature Embedding}
\begin{figure}
\begin{center}
\includegraphics[width=1\linewidth]{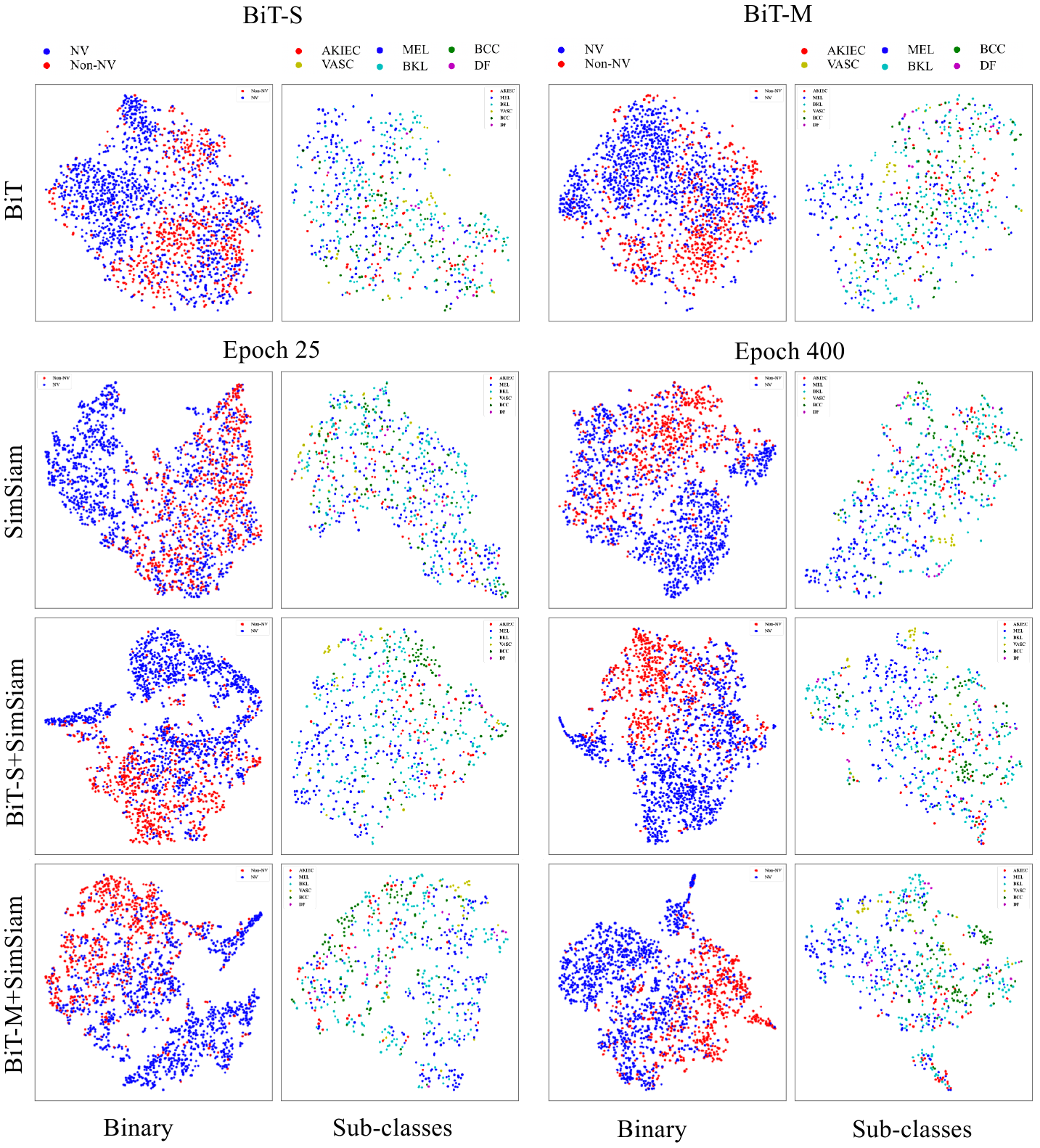}
\end{center}
   \caption{\textbf{The t-SNE plots of the testing images on HAM10000.} The top panels show the t-SNE plots of directly applying BiT-S and BiT-M on HAM10000's testing images (without fine tuning). The lower panels show the unsupervised embedding of different approaches on the same testing images at 25th and 400th epoch, respectively. ``Binary" indicates the binary classes (NV vs. Non-NV), while the ``Sub-classes" represents the six sub classes in Non-NV group. }
\label{fig:ham_tsne}
\end{figure}

Fig.~\ref{fig:cifar_tsne} shows the t-SNE plots of SimSiam using Batch Norm and Group Norm, respectively. This plot presents the results in Fig.~\ref{fig:collapse}. Fig.~\ref{fig:ham_tsne} shows the t-SNE plots of BiT-S, BiT-M, SimSiam, BiT-S+SimSiam, and BiT-M+SimSiam on HAM10000 at epoch 25 and 400, respectively. The t-SNE plots of BiT-S and BiT-M are provided as well. Note that the embedding plots are computed on the testing images directly, without applying task specific fine-tuning. The "Binary" scatter plots shows the embedding of all samples with binary classes (NV vs. Non-NV). The "Sub-classes" plots shows the embedding of sub-classes of Non-NV. Generally, the BiT-S+SimSiam and BiT-M+SimSiam achieve more visually separable embedding than BiT-S and BiT-M, in the "Binary" scatter plots. The embedding of both BiT-S+SimSiam and BiT-M+SimSiam at 400th epoch is more visually separable than the 25th epoch in the "Sub-classes" scatter plots. 





\section{Discussion}
\indent\indent\textbf{Model Collapse with Group Norm.}
From the results, the key lesson of adapting BiT to SimSiam is the model collapse using Group Norm. The results show that Batch Norm is essential for SimSiam. This results matches the illustrations in ~\cite{chen2020exploring}, which interpreted the SimSiam as a process with (1) iterative unsupervised clustering and (2) feature updates based on clustering (similar to K-means\cite{likas2003global} or EM\cite{mclachlan2007algorithm} algorithms). This gap can be bridged by replacing the Group Norm layers in BiT with Batch Norm layers, for the downstream SimSiam learning.

\textbf{BiT, SimSiam, and BiT+SimSiam.}
The results demonstrate that BiT is able to speed up the contrastive learning using SimSiam on both CIFAR-10 and HAM10000 datasets. Moreover, the overall performance of BiT-S+SimSiam and BiT-M+SimSiam is better than training SimSiam from scratch. The BiT-M+SimSiam and BiT-S+SimSiam have similar performance, which demonstrates that the model pretrained on a larger size of data does not have noticeable differences for downstream contrastive learning. 

\textbf{Limitations of our works.}
Currently, only BiT and SimSiam have been evaluated, given numerous algorithms in transfer learning and self-supervised learning. Moreover, only a single HAM10000 dataset is evaluated in this pilot study without having comprehensive analysis on multiple and larger medical image datasets. The major reason is that our evaluations are limited by an academic level computational resources, via a single 16GB GPU memory card. The purpose of this study is to motivate researchers to develop a more principle approach of combining big pretrained models with contrastive learning models for image analysis. We hope such strategy would take advantage of big pretrain models with contrastive learning to provide better representation learning.




%
%
\bibliographystyle{splncs04}
\bibliography{main}

\begin{thebibliography}{10}
\providecommand{\url}[1]{\texttt{#1}}
\providecommand{\urlprefix}{URL }
\providecommand{\doi}[1]{https://doi.org/#1}

\bibitem{alzubaidi2020towards}
Alzubaidi, L., Fadhel, M.A., Al-Shamma, O., Zhang, J., Santamar{\'\i}a, J.,
  Duan, Y., Oleiwi, S.R.: Towards a better understanding of transfer learning
  for medical imaging: a case study. Applied Sciences  \textbf{10}(13), ~4523
  (2020)

\bibitem{chen2020simple}
Chen, T., Kornblith, S., Norouzi, M., Hinton, G.: A simple framework for
  contrastive learning of visual representations (2020)

\bibitem{chen2020exploring}
Chen, X., He, K.: Exploring simple siamese representation learning. arXiv
  preprint arXiv:2011.10566  (2020)

\bibitem{deng2009imagenet}
Deng, J., Dong, W., Socher, R., Li, L.J., Li, K., Fei-Fei, L.: Imagenet: A
  large-scale hierarchical image database. In: 2009 IEEE conference on computer
  vision and pattern recognition. pp. 248--255. Ieee (2009)

\bibitem{goyal2017accurate}
Goyal, P., Doll{\'a}r, P., Girshick, R., Noordhuis, P., Wesolowski, L., Kyrola,
  A., Tulloch, A., Jia, Y., He, K.: Accurate, large minibatch sgd: Training
  imagenet in 1 hour. arXiv preprint arXiv:1706.02677  (2017)

\bibitem{grill2020bootstrap}
Grill, J.B., Strub, F., Altch{\'e}, F., Tallec, C., Richemond, P.H.,
  Buchatskaya, E., Doersch, C., Pires, B.A., Guo, Z.D., Azar, M.G., et~al.:
  Bootstrap your own latent: A new approach to self-supervised learning. arXiv
  preprint arXiv:2006.07733  (2020)

\bibitem{he2020momentum}
He, K., Fan, H., Wu, Y., Xie, S., Girshick, R.: Momentum contrast for
  unsupervised visual representation learning. In: Proceedings of the IEEE/CVF
  Conference on Computer Vision and Pattern Recognition. pp. 9729--9738 (2020)

\bibitem{he2016deep}
He, K., Zhang, X., Ren, S., Sun, J.: Deep residual learning for image
  recognition. In: Proceedings of the IEEE conference on computer vision and
  pattern recognition. pp. 770--778 (2016)

\bibitem{he2016identity}
He, K., Zhang, X., Ren, S., Sun, J.: Identity mappings in deep residual
  networks. In: European conference on computer vision. pp. 630--645. Springer
  (2016)

\bibitem{hjelm2018learning}
Hjelm, R.D., Fedorov, A., Lavoie-Marchildon, S., Grewal, K., Bachman, P.,
  Trischler, A., Bengio, Y.: Learning deep representations by mutual
  information estimation and maximization. arXiv preprint arXiv:1808.06670
  (2018)

\bibitem{hosny2018artificial}
Hosny, A., Parmar, C., Quackenbush, J., Schwartz, L.H., Aerts, H.J.: Artificial
  intelligence in radiology. Nature Reviews Cancer  \textbf{18}(8),  500--510
  (2018)

\bibitem{huo2021ai}
Huo, Y., Deng, R., Liu, Q., Fogo, A.B., Yang, H.: Ai applications in renal
  pathology. Kidney International  (2021)

\bibitem{ioffe2015batch}
Ioffe, S., Szegedy, C.: Batch normalization: Accelerating deep network training
  by reducing internal covariate shift. In: International Conference on Machine
  Learning. pp. 448--456 (2015)

\bibitem{kolesnikov2019big}
Kolesnikov, A., Beyer, L., Zhai, X., Puigcerver, J., Yung, J., Gelly, S.,
  Houlsby, N.: Big transfer (bit): General visual representation learning.
  arXiv preprint arXiv:1912.11370  (2019)

\bibitem{cifar100}
Krizhevsky, A.: Learning multiple layers of features from tiny images. Tech.
  rep. (2009)

\bibitem{likas2003global}
Likas, A., Vlassis, N., Verbeek, J.J.: The global k-means clustering algorithm.
  Pattern recognition  \textbf{36}(2),  451--461 (2003)

\bibitem{mclachlan2007algorithm}
McLachlan, G.J., Krishnan, T.: The EM algorithm and extensions, vol.~382. John
  Wiley \& Sons (2007)

\bibitem{micikevicius2018mixed}
Micikevicius, P., Narang, S., Alben, J., Diamos, G., Elsen, E., Garcia, D.,
  Ginsburg, B., Houston, M., Kuchaiev, O., Venkatesh, G., Wu, H.: Mixed
  precision training (2018)

\bibitem{mustafa2021supervised}
Mustafa, B., Loh, A., Freyberg, J., MacWilliams, P., Karthikesalingam, A.,
  Houlsby, N., Natarajan, V.: Supervised transfer learning at scale for medical
  imaging. arXiv preprint arXiv:2101.05913  (2021)

\bibitem{noroozi2016unsupervised}
Noroozi, M., Favaro, P.: Unsupervised learning of visual representations by
  solving jigsaw puzzles. In: European conference on computer vision. pp.
  69--84. Springer (2016)

\bibitem{qiao2019weight}
Qiao, S., Wang, H., Liu, C., Shen, W., Yuille, A.: Weight standardization.
  arXiv preprint arXiv:1903.10520  (2019)

\bibitem{raghu2019transfusion}
Raghu, M., Zhang, C., Kleinberg, J., Bengio, S.: Transfusion: Understanding
  transfer learning for medical imaging. In: Advances in Neural Information
  Processing Systems. pp. 3347--3357 (2019)

\bibitem{razzak2018deep}
Razzak, M.I., Naz, S., Zaib, A.: Deep learning for medical image processing:
  Overview, challenges and the future. Classification in BioApps pp. 323--350
  (2018)

\bibitem{russakovsky2015imagenet}
Russakovsky, O., Deng, J., Su, H., Krause, J., Satheesh, S., Ma, S., Huang, Z.,
  Karpathy, A., Khosla, A., Bernstein, M., et~al.: Imagenet large scale visual
  recognition challenge. International journal of computer vision
  \textbf{115}(3),  211--252 (2015)

\bibitem{shen2017deep}
Shen, D., Wu, G., Suk, H.I.: Deep learning in medical image analysis. Annual
  review of biomedical engineering  \textbf{19},  221--248 (2017)

\bibitem{tschandl2018ham10000}
Tschandl, P., Rosendahl, C., Kittler, H.: The ham10000 dataset, a large
  collection of multi-source dermatoscopic images of common pigmented skin
  lesions. Scientific data  \textbf{5}(1), ~1--9 (2018)

\bibitem{willemink2020preparing}
Willemink, M.J., Koszek, W.A., Hardell, C., Wu, J., Fleischmann, D., Harvey,
  H., Folio, L.R., Summers, R.M., Rubin, D.L., Lungren, M.P.: Preparing medical
  imaging data for machine learning. Radiology  \textbf{295}(1),  4--15 (2020)

\bibitem{wu2018group}
Wu, Y., He, K.: Group normalization. In: Proceedings of the European conference
  on computer vision (ECCV). pp. 3--19 (2018)

\bibitem{wu2018unsupervised}
Wu, Z., Xiong, Y., Yu, S.X., Lin, D.: Unsupervised feature learning via
  non-parametric instance discrimination. In: Proceedings of the IEEE
  Conference on Computer Vision and Pattern Recognition. pp. 3733--3742 (2018)

\bibitem{zhuang2019local}
Zhuang, C., Zhai, A.L., Yamins, D.: Local aggregation for unsupervised learning
  of visual embeddings. In: Proceedings of the IEEE/CVF International
  Conference on Computer Vision. pp. 6002--6012 (2019)

\end{thebibliography}
%




\end{document}